\PassOptionsToPackage{table}{xcolor}
\documentclass[onecolumn,numbers]{main}
\usepackage{amsmath,amssymb,amsfonts}
\microtypesetup{final}
\usepackage{graphicx}
\usepackage{caption}
\usepackage{ragged2e}
\newcommand{\panelref}[1]{\textbf{\textit{(#1)}}}
\titlespacing*{\section}{0pt}{2.4ex plus .5ex minus .3ex}{1.1ex plus .2ex}
\titlespacing*{\subsection}{0pt}{1.9ex plus .4ex minus .2ex}{.8ex plus .2ex}
\titlespacing*{\subsubsection}{0pt}{1.4ex plus .3ex minus .2ex}{.6ex plus .2ex}
\usepackage{wrapfig}
\usepackage{needspace}
\AtBeginEnvironment{figure}{\setlength{\parfillskip}{0pt plus 1fil}}
\AtBeginEnvironment{table}{\setlength{\parfillskip}{0pt plus 1fil}}
\AtBeginEnvironment{wrapfigure}{\setlength{\parfillskip}{0pt plus 1fil}}
\newcommand{\finishwrap}{%
  \par
  \ifnum\value{WF@wrappedlines}>1
    \vspace{\dimexpr\value{WF@wrappedlines}\baselineskip-\baselineskip\relax}%
  \fi
  \WFclear
}
\usepackage{float}
\usepackage{tikz}
\usetikzlibrary{arrows.meta,positioning,calc}
\usepackage{placeins}
\usepackage{booktabs}
\usepackage{multirow}
\usepackage{xcolor}
\usepackage{url}
\definecolor{tarsacpurple}{HTML}{8557A0}
\hypersetup{colorlinks=true,linkcolor=tarsacpurple,citecolor=tarsacpurple,urlcolor=tarsacpurple}
\crefname{table}{Table}{Tables}
\crefname{figure}{Figure}{Figures}
\crefname{section}{Section}{Sections}
\crefname{equation}{Eq.}{Eqs.}
\usepackage{listings}
\usepackage{colortbl}
\makeatletter
\newcommand{\tablecaption}{\def\@captype{table}\caption}
\makeatother

\lstdefinestyle{actionspace}{
    basicstyle=\ttfamily\tiny,
    frame=none,
    breaklines=true,
    breakatwhitespace=false,
    columns=fullflexible,
    keepspaces=true,
    showstringspaces=false,
    xleftmargin=0pt,
    xrightmargin=0pt,
    aboveskip=0pt,
    belowskip=0pt,
    lineskip=-0.4pt
}

\usepackage{rotating}
\usepackage{array}
\newcolumntype{C}[1]{>{\centering\arraybackslash}p{#1}}
\newcolumntype{L}[1]{>{\raggedright\arraybackslash}p{#1}}
\newcommand{\makecell}[1]{\begin{tabular}[b]{@{}c@{}}#1\end{tabular}}
\usepackage{amssymb}

\definecolor{tableaccent}{HTML}{F3EBFA}
\definecolor{tablegray}{RGB}{232,232,232}
\makeatletter
\newsavebox{\tablegroupstrut}
\newcommand{\compacttabgroup}[2]{%
  \noalign{\global\setbox\tablegroupstrut=\copy\@arstrutbox
    \global\setbox\@arstrutbox=\hbox{\vrule
      height .82\ht\tablegroupstrut depth .70\dp\tablegroupstrut width 0pt}}%
  \rowcolor{tablegray}\multicolumn{#1}{@{}l}{\textit{#2}}\\
  \noalign{\global\setbox\@arstrutbox=\copy\tablegroupstrut}%
}
\makeatother
\definecolor{panelgray}{RGB}{240,240,240}

\newcommand{\insight}[1]{
  \par\smallskip
  \begingroup
  \setlength{\fboxsep}{6pt}
  \noindent\colorbox{tarsacpurple!7}{\parbox{\dimexpr\linewidth-2\fboxsep\relax}{
    \textcolor{tarsacpurple}{\textbf{Insight.}}\enspace#1
  }}\par
  \endgroup
  \smallskip
}
\newcommand{\challenge}[3]{
  \noindent\textcolor{tarsacpurple}{\textbf{Challenge~#1: #2}\enspace[\cref{#3}].}
}
\newcommand{\yes}{\checkmark}
\newcommand{\no}{\textcolor{black!35}{\ensuremath{\times}}}
\newcommand{\half}{\textcolor{black!55}{\ensuremath{\bullet}}}

\graphicspath{{fig/}}

\newcommand{\sys}{TARSphere}
\newcommand{\infra}{\sys-Infra}
\newcommand{\bench}{\sys-Bench}
\newcommand{\agent}{\sys-Harness}
\newcommand{\agentFramework}{\sys{} Framework}

\newcommand{\benchmarkBehavior}{BEHAVIOR-1K}
\newcommand{\benchmarkPartnr}{PARTNR}
\newcommand{\benchmarkEmbodiedBench}{EmbodiedBench}
\newcommand{\hyEmbodied}{HY-Embodied}

\newcommand{\humanClaw}{HumanCLAW}
\newcommand{\benchmarkHumanClaw}{\humanClaw-Bench}
\newcommand{\virtualHome}{VirtualHome}
\newcommand{\habitat}{Habitat}
\newcommand{\habitatTwo}{\habitat{} 2.0}
\newcommand{\habitatThree}{\habitat{} 3.0}
\newcommand{\roboThor}{RoboTHOR}
\newcommand{\roboCasa}{RoboCasa}
\newcommand{\maniSkill}{ManiSkill3}
\newcommand{\grScenes}{GRScenes-100}

\newcommand{\sayCan}{SayCan}
\newcommand{\codeAsPolicies}{Code as Policies}
\newcommand{\innerMonologue}{Inner Monologue}
\newcommand{\sayPlan}{SayPlan}
\newcommand{\voxPoser}{VoxPoser}
\newcommand{\rtTwo}{RT-2}
\newcommand{\openVla}{OpenVLA}
\newcommand{\groot}{GR00T}
\newcommand{\ros}{ROS}
\newcommand{\rosTwo}{\ros{} 2}
\newcommand{\isaacSim}{Isaac Sim}
\newcommand{\physX}{PhysX}

\newcommand{\benchmarkAlfworld}{ALFWorld}
\newcommand{\lotaBench}{LoTa-Bench}

\newcommand{\okRobot}{OK-Robot}

\newcommand{\harnessVla}{Harness VLA}
\newcommand{\roboClaw}{RoboClaw}
\newcommand{\aspire}{ASPIRE}
\newcommand{\zetta}{Zetta}

\newcommand{\modelGpt}{GPT-5.6-Terra}
\newcommand{\modelSonnet}{Claude Sonnet-5}
\newcommand{\modelGemini}{Gemini-3.8-Flash}
\newcommand{\modelQwen}{Qwen3.7-Plus}

\newcommand{\modelQwenOpen}{Qwen3.8-27B}
\newcommand{\modelHyEmbodied}{Hy-Embodied-VLM-1.0}
\newcommand{\modelRoboBrain}{RoboBrain2.0-32B}
\newcommand{\modelGemma}{Gemma-4-31B}

\newcommand{\problem}{physically and socially grounded long-horizon embodied
  execution}

\newcommand{\mainnav}{navigate-and-report}
\newcommand{\mainfetch}{fetch-and-deliver}

\newcommand{\slicepr}{perceptual robustness}
\newcommand{\sliceda}{disruption adaptation}
\newcommand{\slicetu}{task update}
\newcommand{\slicehc}{human coordination}
\newcommand{\sliceiu}{intent understanding}

\newcommand{\nscene}{30}
\newcommand{\nepisode}{360}
\newcommand{\nmodel}{8}

\newcommand{\checkTranslationN}{120}
\newcommand{\checkRotationN}{60}
\newcommand{\checkCompositeN}{60}
\newcommand{\checkGraspN}{60}
\newcommand{\checkGraspNearPass}{58}
\newcommand{\checkGraspMidPass}{57}
\newcommand{\checkGraspFarPass}{58}
\newcommand{\checkGraspOuterPass}{56}
\newcommand{\checkGraspTotalPass}{229}
\newcommand{\checkGraspTotalN}{240}
\newcommand{\checkGraspMinRate}{93.3}
\newcommand{\checkGraspMaxRate}{96.7}
\newcommand{\checkGroundN}{300}
\newcommand{\checkDetectPass}{285}
\newcommand{\checkLocalizePass}{260}
\newcommand{\checkLocalizeTolerance}{10}
\newcommand{\checkCurationN}{360}
\newcommand{\checkCurationAgent}{45.0}
\newcommand{\checkCurationHuman}{65.0}
\newcommand{\checkCurationSkill}{72.5}
\newcommand{\checkCurationSimulation}{83.3}
\newcommand{\checkCurationFull}{96.7}
\newcommand{\checkCurationReviewDrop}{51.7}
\newcommand{\haEpisodeN}{180}
\newcommand{\haPlanN}{180}
\newcommand{\haReportN}{180}

\newcommand{\haPlanRho}{0.86}
\newcommand{\haReportRho}{0.91}
\newcommand{\haPlanPlcc}{0.88}
\newcommand{\haReportPlcc}{0.92}

\newcommand{\haHcTaskN}{60}
\newcommand{\haIuTaskN}{60}

\AtBeginEnvironment{thebibliography}{%
  \let\originalbibitem\bibitem
  \renewcommand{\bibitem}[2][]{%
    \Needspace{3\baselineskip}%
    \originalbibitem[#1]{#2}%
    \justifying
    \ifstrequal{#2}{shridhar2020alfworld}{\raggedright}{}%
    \ifstrequal{#2}{qwen38}{\raggedright}{}%
    \ifstrequal{#2}{liu2024ok}{\raggedright}{}%
  }%
}

\renewcommand{\title}[1]{\def\titlelist{{\fontsize{18pt}{22pt}\selectfont\sffamily\bfseries\raggedright #1\par}}}
\title{\sys{}: A Robot-Oriented Full-Stack Framework for Long-Horizon Embodied Agents in Evolving Physical and Social Environments}
\author[1]{Yicheng~Xu}
\author[1]{Zhikai~Xu}
\author[2]{Ziyuan~Chen}
\author[1]{Yuhang~Dong}
\author[1]{Tao~Zhang}
\author[1]{Zhucun~Xue}
\author[1]{Yong~Liu}
\author[1,\dagger]{Jiangning~Zhang}
\affiliation[1]{Zhejiang University}
\affiliation[2]{Fudan University}
\contribution[\dagger]{Corresponding author}
\abstract{
Robots must act under physical and social constraints while adapting to changes in their environment and human needs.
Existing embodied-agent systems often focus on specified goals or abstract execution into symbolic actions, limiting evaluation in an evolving world.
We introduce \sys{}, \textit{a robot-oriented full-stack framework} for physically and socially grounded long-horizon embodied agents.
\infra{} runs robot policies in a physics simulator with embodied NPCs and constrained dialogue.
An event-driven finite-state machine (FSM) uses completed actions and delivered information to update task states and evaluate outcomes.
\agent{}, its embodied brain harness, lets a frozen multimodal large language model (MLLM) act through multi-level executable skills, using observations, hierarchical memory, and execution feedback to revise its decisions.
On this foundation, agent-in-the-loop curation aligns task semantics, causal dependencies, and scene geometry to construct \bench{}: \nepisode{} episodes across \nscene{} indoor scenes.
Five paired conditions assess \slicepr{}, \sliceda{}, \slicetu{}, \slicehc{}, and \sliceiu{} for \mainnav{} and \mainfetch{} tasks.
Our evaluation of \nmodel{} MLLMs shows that general-purpose models can outperform embodied models, while \textit{task updates and human interaction remain difficult}.
Frequent collisions and falls further motivate evaluating execution safety alongside task success, even with a shared action interface and decision budget.
}

\coverdate{October 22, 2026}
\coversourcecode{https://github.com/xuyicheng-zju/TARSphere}
\coverproject{https://xuyicheng-zju.github.io/TARSphere/}
\covercorrespondence{\email{186368@zju.edu.cn}}
\hypersetup{
  pdftitle={TARSphere: A Robot-Oriented Full-Stack Framework for Long-Horizon Embodied Agents in Evolving Physical and Social Environments},
  pdfauthor={Yicheng Xu, Zhikai Xu, Ziyuan Chen, Yuhang Dong, Tao Zhang, Zhucun Xue, Yong Liu, Jiangning Zhang}
}

\begin{document}
\raggedbottom
\setcounter{topnumber}{1}
\setcounter{bottomnumber}{2}
\setcounter{totalnumber}{5}
\renewcommand{\topfraction}{0.92}
\renewcommand{\bottomfraction}{0.75}
\renewcommand{\textfraction}{0.08}
\renewcommand{\floatpagefraction}{0.75}
\maketitle
\begin{figure}[H]
\centering
\includegraphics[width=\textwidth,trim=2bp 2bp 2.5bp 2.5bp,clip]{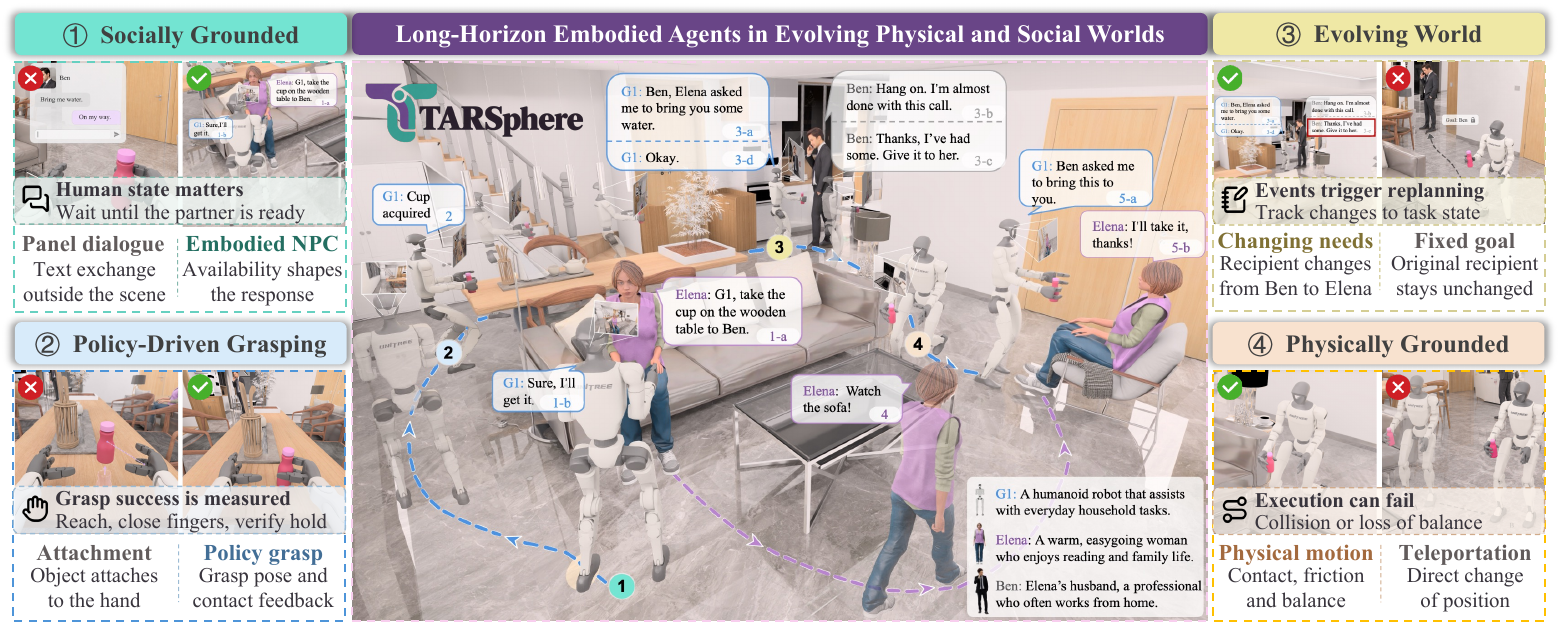}
\caption{\textbf{Robot execution and human interaction in a changing world.}
\panelref{1} Spatially grounded NPC communication depends on human availability and interaction range.
\panelref{2} Policy-driven grasping uses contact feedback to verify a successful hold.
\panelref{3} Evolving task requirements, such as recipient changes, trigger replanning.
\panelref{4} Physical motion is governed by contact, friction, and balance, making collisions and falls part of execution evaluation.}
\label{fig:teaser}
\end{figure}

\clearpage
\setlength{\parfillskip}{0pt plus 0.75\textwidth}
\clubpenalty=10000
\widowpenalty=10000
\displaywidowpenalty=10000
\section{Introduction}
\label{sec:intro}

Robots operating in everyday environments must account for physical constraints~\citep{li2023behavior}, human safety, appropriate interaction, and social norms~\citep{mavrogiannis2023core,puig2024habitat}.
As the world evolves, sensory conditions, physical access, and people's actions and needs can change during execution~\citep{huang2022inner,puig2024habitat,huang2023voxposer}.
Long-horizon tasks therefore require tracking world states, interpreting human intentions, and adjusting plans using execution feedback~\citep{wan2022handmethat,chang2025partnr,huang2022inner} (\cref{fig:teaser}).

Vision-language-action (VLA) models~\citep{brohan2023rt,kim2024openvla} and world-action models (WAMs)~\citep{ye2026world} have advanced robot control, 
yet they often lack the ability to adapt to dynamic and uncertain environments.
Large language models (LLMs) provide broad knowledge for reasoning yet lack the ability to perform complex, real-world tasks~\citep{ahn2022can,huang2022inner,huang2023voxposer}.
This complementarity motivates embodied agents that connect reasoning to action~\citep{ahn2022can,huang2022inner,yang2025embodiedbench,li2026humanclaw}.
However, \harnessVla{}~\citep{zhang2026harness}, \roboClaw{}~\citep{li2026roboclaw}, and \humanClaw{}~\citep{li2026humanclaw} focus on tasks with specified goals, leaving an evolving world with human involvement largely unexamined.
Other settings, including \benchmarkEmbodiedBench{}'s high-level tasks~\citep{yang2025embodiedbench}, \lotaBench{}~\citep{choi2024lota}, \benchmarkAlfworld{}'s text environments~\citep{shridhar2020alfworld}, and Embodied Agent Interface~\citep{li2024embodied}, abstract physical execution into discrete symbolic actions.
GRBench~\citep{wang2024grutopia} combines robot execution with virtual NPC dialogue, while \benchmarkPartnr{}~\citep{chang2025partnr} evaluates embodied human-robot collaboration with temporal task constraints. These gaps motivate a physically and socially grounded foundation for benchmarking robotic agents in an evolving world, with task outcomes tied to execution and human interaction.

\insight{Studying embodied intelligence requires \textit{physically and socially grounded execution} and tasks revealing capabilities across varied situations in a changing world (\cref{sec:main_results,sec:controlled_interventions,sec:execution_checks}).}

\challenge{1}{Embodied execution is not symbolic}{sec:main_results,sec:execution_checks}
A valid action call does not guarantee successful execution: the robot must contend with contact, friction, balance, and human responses.
The agent must therefore revise decisions using physical execution feedback.

\challenge{2}{The world changes during execution}{sec:controlled_interventions,sec:failure_analysis}
Physical conditions, human activity, and task requirements can change while a plan unfolds.
The agent must interpret the current situation and reconsider what to do next.

\challenge{3}{Evaluation must assess diverse capabilities}{sec:main_results}
Within this setting, evaluation must reveal how agents handle varied situations across multiple capability dimensions.
Aggregate success alone can mask different capability gaps across task families (\cref{sec:main_results}).

We introduce \sys{}, a robot-oriented full-stack framework for embodied agents under physical and social constraints.
\infra{} couples robot controllers with embodied NPCs in a physics simulator (\textcolor{tarsacpurple}{Challenge~1}).
An event-driven FSM maintains world and task states, while \agent{} combines observations, memory, skills, and execution feedback for replanning (\textcolor{tarsacpurple}{Challenge~2}).
We build \bench{} through agent-in-the-loop curation, pairing \mainnav{} and \mainfetch{} tasks with five capability conditions: \slicepr{}, \sliceda{}, \slicetu{}, \slicehc{}, and \sliceiu{} (\textcolor{tarsacpurple}{Challenge~3}). Paired variants retain their Base scene and task family.

Across \nmodel{} MLLMs, general-purpose models can outperform embodied models, with capability gaps varying by task family (\cref{sec:main_results}).
Planning credit does not ensure successful execution, and collisions and falls remain common (\cref{sec:main_results,sec:execution_checks}).
These findings call for agents that combine general reasoning, adaptation to world changes, and safe execution (\cref{sec:main_results,sec:controlled_interventions}).

\noindent Our contributions are three-fold.
\begin{itemize}
  \item We formulate long-horizon embodied tasks under physical and social constraints, where execution outcomes and world changes continually shape the agent's decisions.
  \item We develop \agentFramework{}, where physical execution and interaction with embodied humans update task states and supply evaluation evidence; an embodied brain harness uses observations, memory, executable skills, and world feedback to guide model decisions.
  \item We introduce agent-in-the-loop curation for \nepisode{} \bench{} episodes in \nscene{} scenes and five paired capability conditions, aligning task semantics, geometry, and execution.
\end{itemize}

\section{Related Work}
\label{sec:related}

\subsection{Embodied Interactive Simulation Environments}
\label{sec:related_environments}

\virtualHome{}~\citep{puig2018virtualhome} represents household activities as executable programs; \habitat{}~\citep{savva2019habitat} and \roboThor{}~\citep{deitke2020robothor} support navigation and indoor interaction.
\habitatTwo{}~\citep{szot2021habitat} adds rearrangement, and \benchmarkBehavior{}~\citep{li2023behavior} broadens household activities and object-state changes.
\roboCasa{}~\citep{nasiriany2024robocasa} provides diverse kitchen scenes and composite tasks, while \maniSkill{}~\citep{tao2024maniskill3} enables GPU-parallel contact-rich manipulation.
Human participation introduces additional demands: \habitatThree{}~\citep{puig2024habitat} studies social navigation and collaborative rearrangement; \benchmarkPartnr{}~\citep{chang2025partnr} evaluates household collaboration under spatial, temporal, and capability constraints.
GRBench~\citep{wang2024grutopia} combines robot execution with virtual NPC dialogue.
\sys{} couples physical robot execution, embodied NPC behavior, and changing task requirements in one event-driven environment.
Communication depends on proximity, visibility, hearing range, and human availability, while completed actions and delivered messages commit task-state changes.
Paired conditions compare adaptation and interaction within matched scenes and task families, relating capability demands to decisions, physical execution, and completion.

\subsection{Embodied Foundation Models}
\label{sec:related_models}

\sayCan{}~\citep{ahn2022can} grounds skill selection in executable behaviors, and \codeAsPolicies{}~\citep{liang2023code} generates programs combining perception, control APIs, and feedback loops.
\innerMonologue{}~\citep{huang2022inner} uses scene descriptions, success feedback, and human interaction to revise decisions.
\sayPlan{}~\citep{rana2023sayplan} uses hierarchical 3D scene graphs and simulator feedback for planning, while \voxPoser{}~\citep{huang2023voxposer} composes observation-grounded 3D value maps for closed-loop manipulation.
Complementary action models, including \rtTwo{}~\citep{brohan2023rt}, \openVla{}~\citep{kim2024openvla}, and \groot{}~\citep{bjorck2025gr00t}, connect multimodal inputs to robot control; \hyEmbodied{}~\citep{hy2026hy} targets spatial understanding and embodied action.
\sys{} evaluates frozen multimodal models through a shared harness exposing bounded primitives, higher-level skills, hierarchical memory, and measured execution feedback.
This setting examines how model decisions respond to evolving goals, access, and human needs, alongside ablations of the perception, memory, and skill interface supporting those decisions.

\subsection{Robotic Agent Systems and Interfaces}
\label{sec:related_agents}

\ros{}~\citep{quigley2009ros} and \rosTwo{}~\citep{macenski2022robot} provide modular infrastructure for robot perception, planning, and control.
\okRobot{}~\citep{liu2024ok} integrates open-vocabulary perception, navigation, and manipulation for household pick-and-drop tasks.
\roboClaw{}~\citep{li2026roboclaw} links data collection, policy learning, and deployment through a VLM controller orchestrating learned primitives.
\aspire{}~\citep{lu2026aspire} repairs control programs and accumulates reusable skills, while \zetta{}~\citep{ding2026zetta} develops runtime critics and recovery skills around a frozen policy.
To separate decisions from execution, \humanClaw{}~\citep{li2026humanclaw} studies VLM decisions through a simulated human body excluding execution-side disturbances.
\sys{} retains robot dynamics and records collisions, falls, motion residuals, and task outcomes.
Its event-driven evaluator checks evidence against the requirements active when it was produced, including authorized task revisions and recipient acceptance, supporting analysis of how physical execution, adaptation, and human interaction jointly affect completion.

\section{Methodology}
\label{sec:method}

\subsection{Problem Formulation}
\label{sec:problem}

We study \problem{} in a changing simulated world.
The agent interprets human intent from egocentric observations, communication, and interaction history, then pursues the required physical outcome or accurate report.

At time $t_k$, policy $\pi$ selects action sequence $B_k$ from observation $o_{t_k}$ and interaction history $h_{t_k}$:
\begin{equation}
\label{eq:world}
B_k = \pi(o_{t_k}, h_{t_k}),
\qquad
\langle s_{t+1}, f_t \rangle
= \mathcal{W}(s_t, \operatorname{Exec}(a_t), e_t),
\quad a_t \in B_k.
\end{equation}
Here $s_t$ comprises robot, object, human, and active task states; $e_t$ is a human or environmental event.
$\operatorname{Exec}$ invokes the backend; $\mathcal{W}$ updates the world and returns measured feedback $f_t$.
Actions execute sequentially; new observations and feedback inform the next decision.
\agentFramework{} implements this loop; \bench{} supplies tasks and controlled capability conditions there.

\subsection{\agentFramework{}}
\label{sec:framework}

\begin{figure}[t]
\centering
\includegraphics[width=\textwidth]{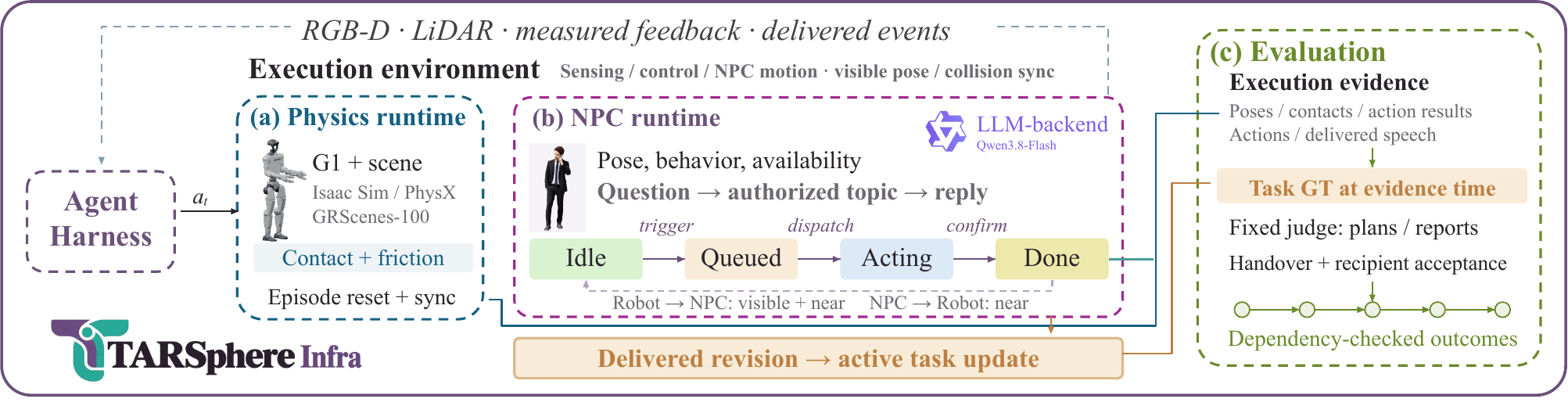}
\caption{\textbf{\infra{}: execution infrastructure.}
\panelref{a} Physics runtime executes robot actions under contact and friction.
\panelref{b} An event-driven FSM coordinates NPC actions, message delivery, and task-state transitions.
\panelref{c} Evaluation checks execution evidence against the task requirements active when that evidence was produced.}
\label{fig:infra}
\end{figure}

\begin{figure}[t]
\centering
\includegraphics[width=\textwidth]{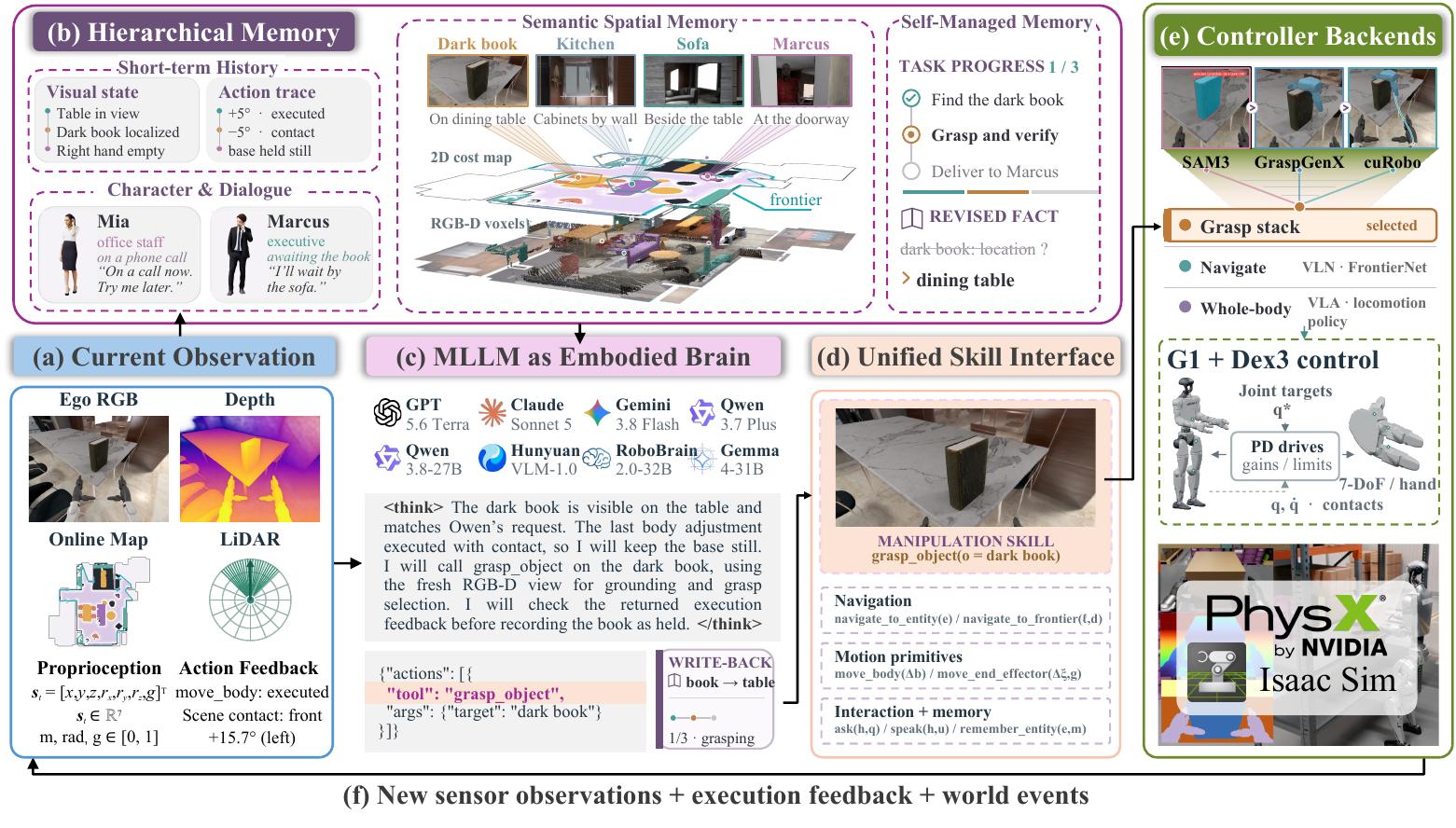}
\caption{\textbf{\agent{}: the embodied brain harness.}
\panelref{a} Current observations and \panelref{b} hierarchical memory inform \panelref{c} MLLM decisions.
\panelref{d} The unified skill interface routes actions to \panelref{e} controller backends.
\panelref{f} Observations, execution feedback, and public world events close the loop.}
\label{fig:harness}
\end{figure}

\agentFramework{} has two layers: \infra{}, the Execution Infrastructure (\cref{fig:infra}), manages physical execution, embodied human interaction, task states, and evaluation.
\agent{}, the Embodied Brain Harness (\cref{fig:harness}), organizes observations, memory, executable skills, and feedback for a frozen MLLM.
\infra{} returns synchronized RGB-D and LiDAR observations, measured feedback, and delivered public events to the harness; task requirements and scoring are private.
We separately test execution fidelity and harness components (\cref{sec:execution_checks,sec:harness_ablations}).

\subsubsection{\infra{}}
\label{sec:infrastructure}

\noindent\textbf{Simulation and robot execution.}
In \cref{fig:infra}, the harness sends robot actions to a shared execution environment containing G1, task objects, and NPCs in \grScenes{}~\citep{wang2024grutopia} scenes.
\isaacSim{}~\citep{NVIDIA_Isaac_Sim} and \physX{} execute robot control under contact and friction: locomotion follows a learned policy, while grasping executes planned arm motion and requires physical object acquisition and retention (\cref{sec:execution_checks}).
Episode reset and synchronization keep sensing, robot control, and NPC motion consistent.
Kinematic motion and skeletal animation keep NPCs' visible poses aligned with their collision proxies, so visible human motion and robot contact refer to the same embodied person throughout execution and interaction.

\noindent\textbf{Embodied NPCs and constrained dialogue.}
The NPC runtime maintains each person's pose, behavior, availability, and authorized knowledge.
Its LLM backend routes natural questions to authorized topics and returns episode-defined replies.
Robot-to-NPC queries require proximity and visibility; NPC speech reaches the robot only within hearing range, so available information depends on the current spatial configuration.

\noindent\textbf{Event-driven task state machine.}
The FSM (\cref{fig:infra}) tracks event and NPC action status and active task bindings.
It admits events when DAG prerequisites and live physical and interaction conditions hold; NPC actions follow \emph{Idle--Queued--Acting--Done}.
\textit{Confirmed action completion and actual message delivery} commit state transitions, applying authorized updates and enabling dependent stages.
The evaluator scores five dependency-checked outcomes against the applicable requirements; task-update events are unscored.
A fixed semantic judge scores plans and reports; handover requires physical execution and recipient acceptance.
We compare semantic scores with human judgments (\cref{sec:human_alignment}) and trace responses to task updates through message delivery, goal revision, and physical completion (\cref{sec:controlled_interventions,sec:failure_analysis}).

\subsubsection{Embodied Brain Harness}
\label{sec:harness}

\noindent\textbf{Closed-loop agent--world interaction.}
Execution outcomes and world changes can invalidate a plan (\cref{sec:controlled_interventions,sec:failure_analysis}).
\agent{} executes actions serially and can return control at tool boundaries, canceling the remaining actions.
Each decision receives current egocentric RGB, body-frame LiDAR, public events, and execution feedback through \cref{fig:harness}(a, f).
Motion residuals and collision bearings report movement error and contact direction for subsequent decisions.
We ablate motion residuals and LiDAR observations to examine their contributions to replanning and task completion (\cref{sec:harness_ablations}).

\noindent\textbf{Multimodal context and hierarchical memory.}
As views and task requirements change, \agent{} retains context through \textit{current observations and the memory layers} in \cref{fig:harness}(a, b), whose contributions we examine through ablations (\cref{sec:harness_ablations}).
\emph{Short-term history} retains recent actions, outcomes, and visual-state notes; \emph{long-term records} retain the active plan, dialogue, and robot state.
\emph{Semantic spatial memory} grounds model-selected people, objects, and places in synchronized RGB-D, storing measured locations and supporting keyframes alongside the online geometric map.
\emph{Self-managed key memory} lets the model retain, revise, or forget selected facts during execution.

\noindent\textbf{Unified multi-level skill interface.}
Through the interface in \cref{fig:harness}(d), the model combines bounded \emph{primitives} for body, gaze, and end-effector motion; \emph{manipulation skills} for locating and grasping visible targets; and \emph{navigation skills} for reaching points, visual frontiers, or remembered entities using the online map.
Social actions use the same interface with spatial preconditions.
The controller backends in \cref{fig:harness}(e) can be replaced by vision-language navigation and action policies.
We ablate navigation and grasp skills and separately assess execution accuracy (\cref{sec:harness_ablations,sec:execution_checks}).

\subsection{\bench{}}
\label{sec:benchmark}

\noindent\textbf{Task Design and Benchmark Taxonomy.}
We construct \nepisode{} \bench{} episodes for \mainnav{} and \mainfetch{} in \nscene{} \grScenes{}~\citep{wang2024grutopia} scenes: 10 homes and 20 commercial spaces.
Each Base episode has \textit{five paired conditions} sharing its scene, task family, and five-stage evaluation.
We test \slicepr{} (PR) through sensing changes and \sliceda{} (DA) through temporary blockage or person/tool unavailability while the goal remains fixed.
For \slicetu{} (TU), an authorized event revises the task; \slicehc{} (HC) requires another person's information or onsite work, and \sliceiu{} (IU) requires clarification or contextual inference.
HC seeks help toward a known goal; IU resolves what the person wants. Base comparisons quantify each condition's effects (\cref{sec:main_results}).

\begin{figure}[t]
\centering
\includegraphics[width=\textwidth,trim=6bp 0bp 1bp 0bp,clip]{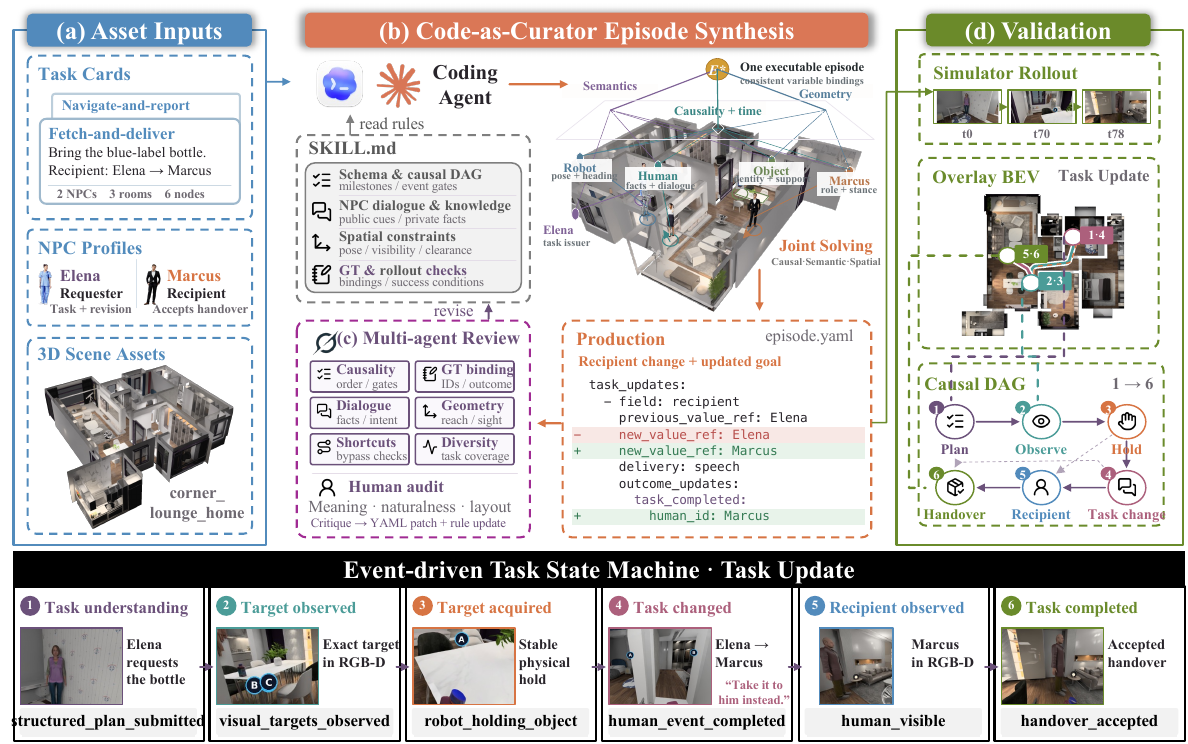}
\caption{\textbf{Agent-in-the-loop curation of \bench{}.}
\panelref{a} Task, NPC, and scene inputs.
\panelref{b} Code-as-curator synthesis with joint semantic, causal, and spatial solving.
\panelref{c} Agent and human review with \texttt{SKILL.md} refinement.
\panelref{d} Simulation validation of task transitions and execution.}
\label{fig:curation}

\end{figure}

\begin{figure}[t]
\centering
\includegraphics[width=\textwidth]{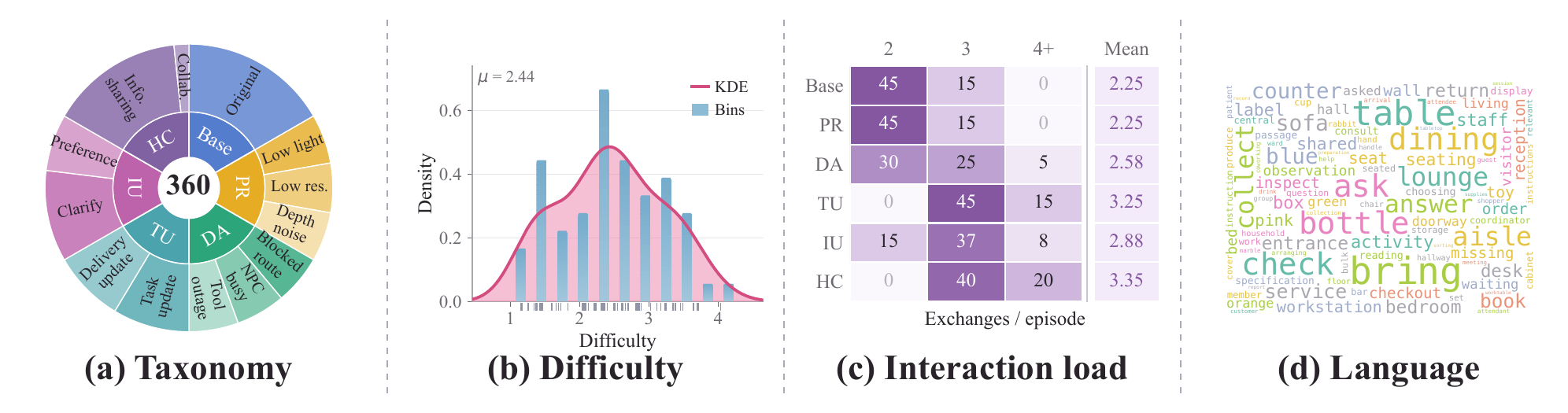}
\caption{\textbf{\bench{} statistics.}
\panelref{a} Base and five conditions across \nepisode{} episodes.
\panelref{b} Difficulty of 60 Base tasks with a kernel density estimate.
\panelref{c} Human exchanges per episode, including the opening; last column: mean.
\panelref{d} Top 100 terms from 240 distinct opening utterances.}
\label{fig:benchmark_statistics}
\par\medskip
\begingroup
\captionsetup{type=table}
\begingroup
\centering
\tablecaption{\textbf{Comparison with existing embodied benchmarks.}}
\label{tab:comparison}
\renewcommand{\arraystretch}{1.08}
\setlength{\tabcolsep}{0.5pt}
\scriptsize
\newlength{\comparisonwidth}
\setlength{\comparisonwidth}{\dimexpr\textwidth-20\tabcolsep\relax}
\resizebox{\textwidth}{!}{%
\begin{tabular}{@{}>{\raggedright\arraybackslash}p{\dimexpr0.17\comparisonwidth-3pt\relax} C{\dimexpr0.05\comparisonwidth-2pt\relax} C{0.0866666667\comparisonwidth} C{\dimexpr0.0866666667\comparisonwidth+1.5pt\relax} C{\dimexpr0.0866666667\comparisonwidth-3.5pt\relax} C{\dimexpr0.0866666667\comparisonwidth-3pt\relax} C{0.0866666667\comparisonwidth} C{\dimexpr0.0866666667\comparisonwidth-2pt\relax} *{2}{C{0.0866666667\comparisonwidth}} C{\dimexpr0.0866666667\comparisonwidth+12pt\relax}@{}}
\toprule
\multicolumn{1}{c}{\multirow{2}{*}[-0.7em]{\textbf{Benchmark}}}
& \multirow{2}{*}[-0.7em]{\textbf{Year}}
& \multicolumn{4}{c}{\textbf{Task Setting}}
& \multicolumn{2}{c}{\textbf{Execution Fidelity}}
& \multicolumn{3}{c}{\textbf{Evaluation Design}} \\
\cmidrule(lr){3-6}\cmidrule(lr){7-8}\cmidrule(lr){9-11}
& & \makecell{Continuous\\Physics} & \makecell{Embodied\\Humans} & \makecell{Dynamic\\Humans} & \makecell{Evolving\\World}
& \makecell{Robot\\Kinematics} & \makecell{Policy\\Driven}
& \makecell{Progressive\\Eval.} & \makecell{Capability\\Contrasts} & \makecell{Event-driven\\Task FSM} \\
\midrule
\benchmarkBehavior{} & 2023 & \half & \no & \no & \no & \yes & \half & \no & \no & \no \\
GRBench & 2024 & \half & \no & \no & \no & \yes & \yes & \no & \no & \no \\
\benchmarkPartnr{} & 2024 & \no & \yes & \yes & \no & \half & \half & \yes & \no & \no \\
\benchmarkEmbodiedBench{} & 2025 & \no & \no & \no & \no & \half & \no & \no & \half & \no \\
\benchmarkHumanClaw{} & 2026 & \half & \no & \no & \no & \no & \yes & \yes & \no & \no \\
\midrule
\rowcolor{tableaccent}
\textbf{\bench{}} & 2026 & \yes & \yes & \yes & \yes & \yes & \yes & \yes & \yes & \yes \\
\bottomrule
\end{tabular}
}
\par\smallskip
{\scriptsize \yes\ supported; \half\ restricted; \no\ unestablished in evaluated execution or scoring.
\par}
\endgroup

\endgroup
\end{figure}

\noindent\textbf{Agent-in-the-Loop Curation.}
Executable task construction is challenging because \textit{semantics, causal dependencies, and scene geometry} must agree.
\Cref{fig:curation} outlines four steps.
\textbf{\textit{(a)~Asset Inputs.}} Task cards, NPC profiles, and scene assets specify intent, authorized NPC knowledge, and candidate objects and locations.
\textbf{\textit{(b)~Code-as-Curator Synthesis and Joint Solving.}} A coding agent follows \texttt{SKILL.md} to jointly solve entity bindings, event dependencies, and interaction positions, producing specifications that align task semantics and state transitions with scene geometry.
\textbf{\textit{(c)~Multi-agent Review and Human Audit.}} Specialized agents check causality, dialogue, outcome bindings, scoring shortcuts, geometry, and diversity.
Human reviewers assess meaning, naturalness, and layout; critiques guide episode corrections and \texttt{SKILL.md} refinement.
\textbf{\textit{(d)~Simulation Validation.}} Simulation checks event delivery and physical execution.
We ablate review, curator refinement, and simulation (\cref{sec:curation_ablations}); human evaluation assesses causal soundness and feasibility (\cref{sec:human_alignment}).
\Cref{fig:benchmark_statistics} summarizes task and interaction statistics.

\section{Experiments}
\label{sec:experiments}

\subsection{Experimental Setup}

We evaluate \nmodel{} frozen MLLMs: four proprietary models, \modelGpt{}, \modelSonnet{}, \modelGemini{}, and \modelQwen{}; two general-purpose open VLMs, \modelQwenOpen{}~\citep{qwen38} and \modelGemma{}~\citep{gemmateam2026gemma4}; and two open embodied models, \modelHyEmbodied{}~\citep{wang2026hy} and \modelRoboBrain{}~\citep{team2025robobrain}.
All use \agent{} on \bench{}'s \nepisode{} episodes, sharing prompts, observations, actions, a 100-decision cap, and termination rules.
By default, decisions use the current image and 10 text-history steps to propose 1-5 actions.
All model weights remain frozen, allowing comparisons through the same harness without benchmark-specific training, parameter updates, or adaptation to the evaluation tasks.

\noindent\textbf{Evaluation Metrics.}
We use quality scoring for the main results and ablations.
The five stages are task understanding, onsite target observation, onsite task completion, recipient observation, and final task completion.
The onsite outcome is an inspection conclusion, an answer received from a source, or a physical grasp, depending on the task.
For episode $e$, let $g_{e,i}\in\{0,1\}$ indicate eligible evidence for stage $i$ under its event dependencies and live conditions, and let $q_{e,i}\in[0,1]$ score that evidence.
Plan, inspection, and delivered-report quality are continuous; observation, information receipt, and physical outcomes are binary.
\par
\begin{samepage}
Let $B_{m,t}^{+}$ contain successful scored Base episodes for model $m$ and task family $t$. We report
\begin{equation}
\mathrm{Prog}(e)=\frac{1}{5}\sum_{i=1}^{5}g_{e,i}q_{e,i},
\quad
\mathrm{Succ}(e)=c_e(1-u_e),
\quad
\mathrm{Eff}_{m,t}=\frac{1}{|B_{m,t}^{+}|}\sum_{e\in B_{m,t}^{+}}\frac{L_{m,e}}{d_e^{\star}}.
\label{eq:outcome_scores}
\end{equation}
Here $c_e$ indicates a correct, complete report to the intended recipient or accepted handover of the correct object, with all prerequisites fulfilled; $u_e$ indicates terminal fall or object loss.
\end{samepage}

Initial plan and inspection quality contribute to Progress; Success depends on the completed task.
Terminal physical failure preserves earned Progress; ordinary contact alone does not veto Success.
Scores average valid runs and are shown as percentages; Avg. equally weights the six conditions.

In the path ratio, $L_{m,e}$ integrates pelvis XY movement at native physics steps, excluding reset and warmup, and $d_e^{\star}$ is the authored Base route length.
Eff. uses successful Base runs with complete paths; unavailable entries are dashes.
We also report collision/fall incidence and decision/token cost over successful and failed runs.

\subsection{Main Results}
\label{sec:main_results}

\begin{table}[t]
\begingroup
\normalsize
\centering
\caption{\textbf{Main results on \bench{}.} Cells: progress/success (\%); Rk.: mean-progress rank per task. Sky-blue bold/rose underlined: best/second-best task scores and Rk. Eff.: successful-Base path length/authored route; Cost: decisions/tokens per decision; Safety: collision/fall episode incidence (\%). Means use valid scored runs. $^{\ast}$/$^{\dagger}$/$^{\ddagger}$: proprietary/general open-weight/embodied open-weight models.}
\label{tab:main}
\renewcommand{\arraystretch}{1.04}
\setlength{\tabcolsep}{0.3pt}
\scriptsize
\newcommand{\pscell}[2]{#1/#2}
\newcommand{\costcell}[1]{\mbox{#1}}
\newcommand{\safetycell}[2]{\mbox{#1/#2}}
\definecolor{rankbest}{HTML}{DDF3FF}
\newcommand{\bestval}[1]{{\setlength{\fboxsep}{0.4pt}\colorbox{rankbest}{\textbf{#1}}}}
\definecolor{ranksecond}{HTML}{FFE4EC}
\newcommand{\runnerupval}[1]{{\setlength{\fboxsep}{0.4pt}\colorbox{ranksecond}{\underline{#1}}}}
\newlength{\mainresultswidth}
\setlength{\mainresultswidth}{\dimexpr\textwidth-22\tabcolsep-2pt\relax}
\resizebox{\textwidth}{!}{%
\begin{tabular}{@{}>{\raggedright\arraybackslash}p{0.20\mainresultswidth} *{7}{>{\centering\arraybackslash}p{0.074\mainresultswidth}} >{\centering\arraybackslash}p{0.04\mainresultswidth} >{\centering\arraybackslash}p{0.055\mainresultswidth} >{\centering\arraybackslash}p{0.097\mainresultswidth} >{\centering\arraybackslash}p{0.09\mainresultswidth}@{}}
\toprule
\multicolumn{1}{c}{\multirow{2}{*}[-0.7em]{\textbf{Model}}}
& \multicolumn{8}{c}{\textbf{Progress / Success by Task Condition} (\%)}
& \multicolumn{3}{c}{\textbf{Execution Quality}} \\
\cmidrule(lr){2-9}\cmidrule(lr){10-12}
& Base $\uparrow$ & PR $\uparrow$ & DA $\uparrow$ & TU $\uparrow$ & HC $\uparrow$ & IU $\uparrow$ & \textbf{Avg. $\uparrow$}
& \mbox{\textbf{Rk.}$\downarrow$} & Eff. $\downarrow$ & Cost $\downarrow$ & Safety $\downarrow$ \\
\midrule
\compacttabgroup{12}{\mainnav{}}
\modelGpt{}$^{\ast}$ & \pscell{56.7}{\runnerupval{23.3}} & \pscell{51.3}{13.3} & \pscell{54.2}{16.7} & \pscell{39.1}{0.0} & \pscell{\runnerupval{55.9}}{\runnerupval{26.7}} & \pscell{45.1}{10.0} & \pscell{50.4}{15.0} & 3 & 1.17 & \costcell{48.8/6.4k} & \safetycell{87}{16} \\
\modelSonnet{}$^{\ast}$ & \pscell{53.9}{\runnerupval{23.3}} & \pscell{\runnerupval{62.0}}{\runnerupval{33.3}} & \pscell{\runnerupval{61.9}}{\runnerupval{31.0}} & \pscell{40.7}{3.3} & \pscell{43.4}{3.4} & \pscell{\bestval{46.0}}{6.7} & \pscell{\runnerupval{51.3}}{\runnerupval{16.9}} & \runnerupval{2} & 1.48 & \costcell{54.9/12.9k} & \safetycell{93}{33} \\
\modelGemini{}$^{\ast}$ & \pscell{\bestval{78.5}}{\bestval{60.0}} & \pscell{\bestval{78.7}}{\bestval{56.7}} & \pscell{\bestval{78.6}}{\bestval{60.0}} & \pscell{\bestval{52.7}}{\bestval{16.7}} & \pscell{\bestval{69.5}}{\bestval{33.3}} & \pscell{\runnerupval{45.2}}{\bestval{16.7}} & \pscell{\bestval{67.2}}{\bestval{40.6}} & \bestval{1} & 1.31 & \costcell{38.9/10.2k} & \safetycell{84}{14} \\
\modelQwen{}$^{\ast}$ & \pscell{\runnerupval{62.0}}{\runnerupval{23.3}} & \pscell{50.9}{13.3} & \pscell{56.6}{16.7} & \pscell{38.7}{0.0} & \pscell{39.5}{10.0} & \pscell{38.0}{3.4} & \pscell{47.6}{11.1} & 5 & 2.45 & \costcell{58.5/7.0k} & \safetycell{97}{34} \\
\cmidrule(lr){1-12}
\modelQwenOpen{}$^{\dagger}$ & \pscell{37.3}{0.0} & \pscell{37.3}{0.0} & \pscell{38.9}{0.0} & \pscell{38.7}{0.0} & \pscell{32.2}{0.0} & \pscell{36.2}{0.0} & \pscell{36.8}{0.0} & 6 & \textemdash & \costcell{17.9/6.3k} & \safetycell{88}{9} \\
\modelGemma{}$^{\dagger}$ & \pscell{50.7}{\runnerupval{23.3}} & \pscell{57.3}{23.3} & \pscell{48.9}{6.7} & \pscell{\runnerupval{42.8}}{\runnerupval{6.9}} & \pscell{43.4}{6.7} & \pscell{45.1}{\runnerupval{13.3}} & \pscell{48.0}{13.4} & 4 & 1.28 & \costcell{59.3/6.7k} & \safetycell{95}{35} \\
\modelRoboBrain{}$^{\ddagger}$ & \pscell{23.2}{0.0} & \pscell{28.1}{0.0} & \pscell{30.6}{0.0} & \pscell{31.4}{0.0} & \pscell{22.4}{0.0} & \pscell{23.7}{0.0} & \pscell{26.6}{0.0} & 7 & \textemdash & \costcell{59.7/6.9k} & \safetycell{98}{46} \\
\modelHyEmbodied{}$^{\ddagger}$ & \pscell{15.9}{0.0} & \pscell{16.0}{0.0} & \pscell{13.9}{0.0} & \pscell{16.1}{0.0} & \pscell{9.1}{0.0} & \pscell{8.6}{0.0} & \pscell{13.2}{0.0} & 8 & \textemdash & \costcell{16.4/6.1k} & \safetycell{91}{19} \\
\midrule
\compacttabgroup{12}{\mainfetch{}}
\modelGpt{}$^{\ast}$ & \pscell{47.0}{16.7} & \pscell{44.8}{13.3} & \pscell{47.1}{16.7} & \pscell{46.0}{6.7} & \pscell{38.4}{6.7} & \pscell{44.3}{16.7} & \pscell{44.6}{12.8} & 3 & 2.38 & \costcell{77.0/6.5k} & \safetycell{98}{24} \\
\modelSonnet{}$^{\ast}$ & \pscell{\runnerupval{51.7}}{\runnerupval{23.3}} & \pscell{\runnerupval{55.1}}{\runnerupval{30.0}} & \pscell{\runnerupval{55.5}}{\runnerupval{30.0}} & \pscell{\runnerupval{46.5}}{\runnerupval{16.7}} & \pscell{\runnerupval{50.2}}{\runnerupval{24.1}} & \pscell{\runnerupval{49.4}}{\runnerupval{23.3}} & \pscell{\runnerupval{51.4}}{\runnerupval{24.6}} & \runnerupval{2} & 2.21 & \costcell{66.1/12.4k} & \safetycell{100}{29} \\
\modelGemini{}$^{\ast}$ & \pscell{\bestval{75.5}}{\bestval{60.0}} & \pscell{\bestval{70.0}}{\bestval{46.7}} & \pscell{\bestval{73.3}}{\bestval{50.0}} & \pscell{\bestval{69.2}}{\bestval{46.7}} & \pscell{\bestval{64.1}}{\bestval{43.3}} & \pscell{\bestval{64.6}}{\bestval{40.0}} & \pscell{\bestval{69.4}}{\bestval{47.8}} & \bestval{1} & 2.06 & \costcell{59.3/10.4k} & \safetycell{98}{16} \\
\modelQwen{}$^{\ast}$ & \pscell{39.6}{10.0} & \pscell{38.1}{3.3} & \pscell{34.8}{3.3} & \pscell{39.7}{6.7} & \pscell{27.1}{0.0} & \pscell{36.3}{6.7} & \pscell{35.9}{5.0} & 4 & 1.93 & \costcell{75.4/6.8k} & \safetycell{100}{34} \\
\cmidrule(lr){1-12}
\modelQwenOpen{}$^{\dagger}$ & \pscell{33.7}{0.0} & \pscell{28.6}{0.0} & \pscell{33.5}{0.0} & \pscell{29.2}{0.0} & \pscell{29.0}{0.0} & \pscell{29.0}{0.0} & \pscell{30.5}{0.0} & 6 & \textemdash & \costcell{58.9/6.5k} & \safetycell{98}{17} \\
\modelGemma{}$^{\dagger}$ & \pscell{36.7}{3.3} & \pscell{34.7}{0.0} & \pscell{37.2}{10.0} & \pscell{32.6}{0.0} & \pscell{27.2}{0.0} & \pscell{30.7}{0.0} & \pscell{33.2}{2.2} & 5 & 2.55 & \costcell{73.3/6.7k} & \safetycell{100}{36} \\
\modelRoboBrain{}$^{\ddagger}$ & \pscell{19.2}{0.0} & \pscell{22.2}{0.0} & \pscell{22.1}{0.0} & \pscell{22.3}{0.0} & \pscell{18.3}{0.0} & \pscell{12.5}{0.0} & \pscell{19.4}{0.0} & 7 & \textemdash & \costcell{61.4/6.6k} & \safetycell{98}{47} \\
\modelHyEmbodied{}$^{\ddagger}$ & \pscell{10.8}{0.0} & \pscell{11.8}{0.0} & \pscell{11.0}{0.0} & \pscell{9.9}{0.0} & \pscell{11.3}{0.0} & \pscell{9.1}{0.0} & \pscell{10.6}{0.0} & 8 & \textemdash & \costcell{22.2/6.2k} & \safetycell{88}{13} \\
\bottomrule
\end{tabular}
}
\par
\endgroup

\end{table}

\textbf{\textit{(1)} Proprietary models lead; the best open VLM outperforms embodied models.}
In \cref{tab:main}, \modelGemini{} achieves 40.6\% navigation and 47.8\% delivery success, compared with 16.9\% and 24.6\% for runner-up \modelSonnet{}.
Under the shared \agent{}, proprietary models occupy the top two ranks in both task families. \textit{This lead suggests that broad multimodal competence may be critical for general embodied tasks.}
Among open models, \modelGemma{} succeeds in both families, while both embodied models have low progress and complete no episodes.
The stage breakdown in \cref{fig:task_progress} shows that planning credit does not ensure target observation or onsite execution.
These results motivate \textit{preserving multimodal abilities in embodied foundation model training} and testing transfer across diverse tasks and human interactions.

\FloatBarrier
\par\medskip\noindent\begin{minipage}{\textwidth}
\begin{wrapfigure}{r}{0.48\textwidth}
\vspace{-\intextsep}
\setlength{\abovecaptionskip}{2pt}
\setlength{\belowcaptionskip}{0pt}
\centering
\includegraphics[width=\linewidth]{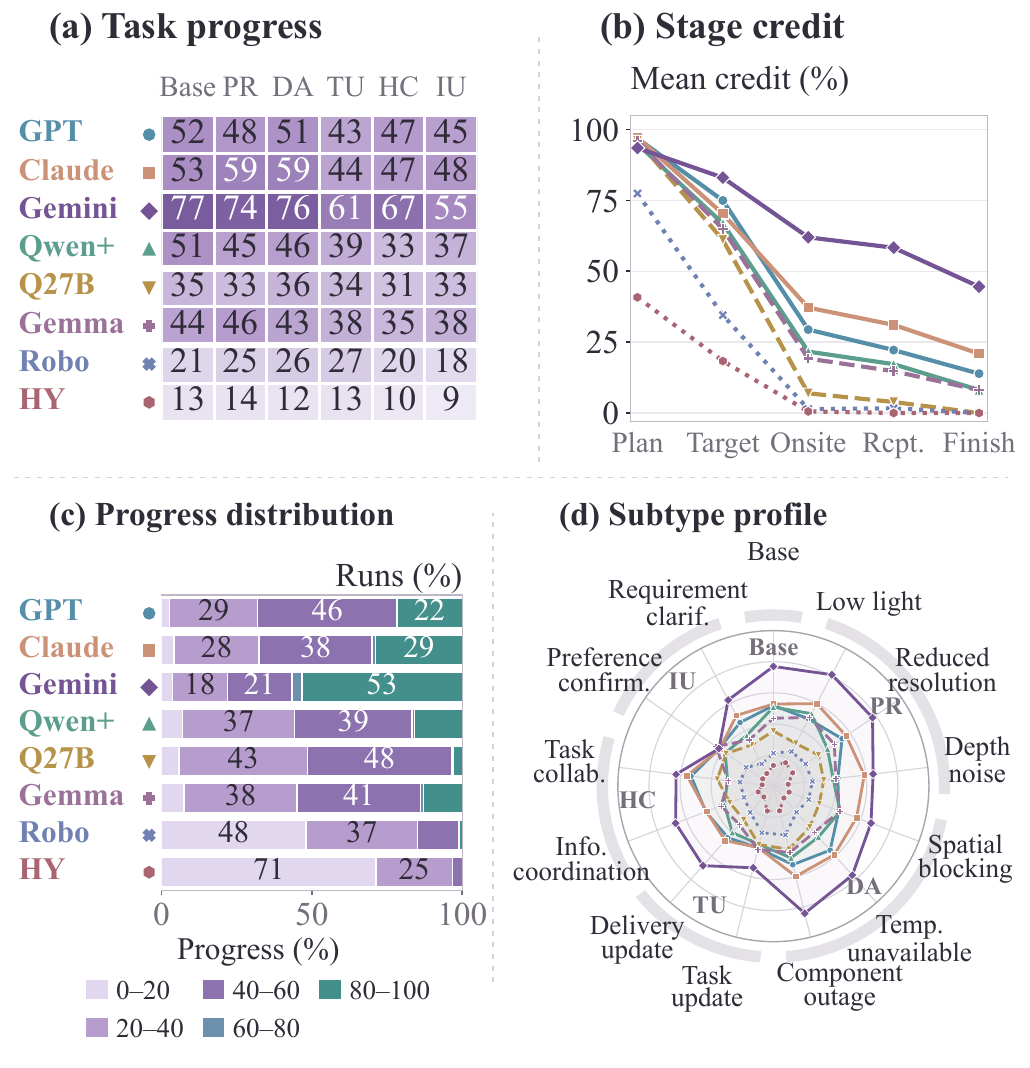}
\caption{\textbf{Task progress.}
\panelref{a} Conditions; \panelref{b} stage credit; \panelref{c} progress bins; \panelref{d} subtypes.}
\label{fig:task_progress}
\end{wrapfigure}

\textbf{\textit{(2)} The taxonomy reveals task-dependent capability gaps.}
Averaged across models, PR and DA retain near-Base Progress in both task families, whereas the largest deficits occur under IU and TU in navigation and HC and IU in delivery (\cref{tab:main}).
\modelGemini{} illustrates the contrasting profiles at identical 60.0\% Base success:
PR/DA success is higher in navigation (56.7-60.0\% vs.\ 46.7-50.0\%), while TU/HC/IU success is higher in delivery (40.0-46.7\% vs.\ 16.7-33.3\%).
Strong performance on perceptual and disruption variants therefore coexists with substantial gaps in task updates, coordination, and intent understanding.
\textit{Similar Base success can therefore conceal different capability gaps} across task families.

\textbf{\textit{(3)} Frequent collisions and falls persist even in the strongest model.}
\modelGemini{} uses fewer decisions than \modelSonnet{} despite its higher success, while \modelGpt{} has the lowest path ratio on successful Base navigation runs at 1.17.
Across model--task groups, collisions occur in 84-100\% of episodes and falls in 9-47\%.
Even \modelGemini{} incurs collisions in 84\% of navigation and 98\% of delivery episodes, with falls in 14\% and 16\%, respectively.
\textit{Completion and efficiency must therefore be assessed alongside physical safety, including collisions and falls throughout execution} (\cref{tab:main}).

\finishwrap
\end{minipage}
\par\medskip

\FloatBarrier
\subsection{Ablation Study and Analysis}

\subsubsection{Harness Ablations}
\label{sec:harness_ablations}

\begin{table}[t]
\begingroup
\normalsize
\centering
\caption{\textbf{Harness ablations.} \modelGemini{} as evaluated model. Experiments are conducted on a carefully selected subset balanced across task families and capability conditions.}
\label{tab:harness_ablation}
\renewcommand{\arraystretch}{1.04}
\setlength{\tabcolsep}{0.5pt}
\scriptsize
\newcommand{\ps}[2]{\mbox{#1/#2}}
\newcommand{\ct}[2]{\mbox{#1/#2k}}
\newcommand{\sfcell}[2]{\mbox{#1/#2}}
\definecolor{rankbest}{HTML}{DDF3FF}
\definecolor{ranksecond}{HTML}{FFE4EC}
\newcommand{\bestval}[1]{{\setlength{\fboxsep}{0.4pt}\colorbox{rankbest}{\textbf{#1}}}}
\newcommand{\runnerupval}[1]{{\setlength{\fboxsep}{0.4pt}\colorbox{ranksecond}{\underline{#1}}}}
\newcommand{\grp}[1]{\compacttabgroup{12}{#1}}
\newlength{\harnessresultswidth}
\setlength{\harnessresultswidth}{\dimexpr\textwidth-22\tabcolsep-2pt\relax}
\resizebox{\textwidth}{!}{%
\begin{tabular}{@{}>{\raggedright\arraybackslash}p{0.075\harnessresultswidth} >{\raggedright\arraybackslash}p{0.165\harnessresultswidth} *{7}{>{\centering\arraybackslash}p{0.0775\harnessresultswidth}} >{\centering\arraybackslash}p{0.055\harnessresultswidth} >{\centering\arraybackslash}p{0.0825\harnessresultswidth} >{\centering\arraybackslash}p{0.08\harnessresultswidth}@{}}
\toprule
\multicolumn{2}{c}{\multirow{2}{*}[-0.7em]{\textbf{Setting}}}
& \multicolumn{7}{c}{\textbf{Task Progress / Success Rate by Condition} (\%)}
& \multicolumn{3}{c}{\textbf{Execution Quality}} \\
\cmidrule(lr){3-9}\cmidrule(lr){10-12}
& & Base $\uparrow$ & PR $\uparrow$ & DA $\uparrow$ & TU $\uparrow$ & HC $\uparrow$ & IU $\uparrow$ & \textbf{Avg. $\uparrow$}
& Eff. $\downarrow$ & Cost $\downarrow$ & Safety $\downarrow$ \\
\midrule
\grp{Mechanisms vs. full harness}
\multirow{2}{*}{\textit{Obs.}} & w/o LiDAR & \ps{72.6}{41.7} & \ps{70.7}{\runnerupval{50.0}} & \ps{54.5}{25.0} & \ps{61.7}{33.3} & \ps{57.8}{25.0} & \ps{48.5}{\bestval{25.0}} & \ps{61.0}{33.3} & 1.12 & \ct{37.1}{9.6} & \sfcell{97}{32} \\ 
& w/o action residuals & \ps{74.2}{50.0} & \ps{75.8}{41.7} & \ps{67.6}{41.7} & \ps{55.0}{16.7} & \ps{59.3}{33.3} & \ps{48.0}{\bestval{25.0}} & \ps{63.3}{34.7} & 1.18 & \ct{49.8}{10.1} & \sfcell{94}{19} \\ 
\addlinespace[1pt]
\multirow{2}{*}{\textit{Memory}} & w/o semantic spatial & \ps{\runnerupval{78.0}}{41.7} & \ps{\runnerupval{76.3}}{\bestval{58.3}} & \ps{69.1}{33.3} & \ps{\runnerupval{68.3}}{\runnerupval{41.7}} & \ps{67.6}{\bestval{58.3}} & \ps{46.8}{16.7} & \ps{67.7}{\runnerupval{41.7}} & 1.40 & \ct{49.5}{9.9} & \sfcell{94}{19} \\ 
& w/o self-managed & \ps{76.2}{50.0} & \ps{73.0}{\runnerupval{50.0}} & \ps{72.3}{25.0} & \ps{65.8}{25.0} & \ps{68.7}{\runnerupval{41.7}} & \ps{48.2}{16.7} & \ps{67.4}{34.7} & 1.00 & \ct{41.0}{9.8} & \sfcell{92}{24} \\ 
\addlinespace[1pt]
\multirow{2}{*}{\textit{Skills}} & w/o navigation skill & \ps{75.7}{50.0} & \ps{67.3}{\runnerupval{50.0}} & \ps{71.2}{41.7} & \ps{65.0}{33.3} & \ps{68.7}{33.3} & \ps{46.0}{8.3} & \ps{65.7}{36.1} & 1.08 & \ct{49.4}{8.0} & \sfcell{90}{15} \\ 
& w/o grasp skill & \ps{55.7}{25.0} & \ps{56.0}{16.7} & \ps{57.4}{33.3} & \ps{53.2}{25.0} & \ps{53.0}{25.0} & \ps{30.7}{0.0} & \ps{51.0}{20.8} & 1.34 & \ct{59.5}{10.6} & \sfcell{90}{19} \\ 
\addlinespace[2pt]
\grp{Configuration vs. full harness}
\multirow{2}{*}{\textit{Text hist.}} & 0 steps & \ps{69.2}{41.7} & \ps{64.0}{33.3} & \ps{\runnerupval{74.4}}{\runnerupval{50.0}} & \ps{65.0}{33.3} & \ps{64.5}{33.3} & \ps{\bestval{53.2}}{\bestval{25.0}} & \ps{65.1}{36.1} & 1.10 & \ct{47.4}{9.9} & \sfcell{94}{19} \\ 
& 50 steps & \ps{71.0}{41.7} & \ps{71.3}{41.7} & \ps{69.2}{33.3} & \ps{\bestval{71.7}}{\runnerupval{41.7}} & \ps{\runnerupval{70.2}}{33.3} & \ps{45.7}{16.7} & \ps{66.5}{34.7} & 1.66 & \ct{44.8}{11.2} & \sfcell{94}{19} \\ 
\addlinespace[1pt]
\multirow{2}{*}{\textit{Vis. hist.}} & 2 frames & \ps{\bestval{82.9}}{\bestval{66.7}} & \ps{\bestval{78.0}}{\runnerupval{50.0}} & \ps{71.1}{33.3} & \ps{61.5}{25.0} & \ps{\bestval{72.3}}{\runnerupval{41.7}} & \ps{49.5}{16.7} & \ps{\runnerupval{69.2}}{38.9} & 1.19 & \ct{47.0}{10.6} & \sfcell{96}{17} \\ 
& 4 frames & \ps{76.1}{50.0} & \ps{64.7}{41.7} & \ps{69.6}{33.3} & \ps{58.3}{33.3} & \ps{57.7}{\runnerupval{41.7}} & \ps{51.5}{\bestval{25.0}} & \ps{63.0}{37.5} & 1.21 & \ct{41.3}{10.5} & \sfcell{93}{22} \\ 
\addlinespace[1pt]
\multirow{2}{*}{\textit{Horizon}} & 3 actions & \ps{68.1}{33.3} & \ps{74.4}{\bestval{58.3}} & \ps{69.8}{41.7} & \ps{65.5}{36.4} & \ps{59.5}{27.3} & \ps{50.1}{\runnerupval{18.2}} & \ps{64.6}{35.9} & 1.38 & \ct{48.2}{10.9} & \sfcell{96}{23} \\ 
& 5 actions & \ps{\runnerupval{78.0}}{50.0} & \ps{73.0}{33.3} & \ps{67.7}{25.0} & \ps{67.3}{36.4} & \ps{66.5}{33.3} & \ps{\runnerupval{52.5}}{16.7} & \ps{67.5}{32.5} & 1.94 & \ct{50.1}{10.9} & \sfcell{96}{21} \\ 
\midrule
\rowcolor{tableaccent}
\multicolumn{2}{@{}l}{\agent{}} & \ps{\runnerupval{78.0}}{\runnerupval{58.3}} & \ps{\runnerupval{76.3}}{41.7} & \ps{\bestval{77.4}}{\bestval{58.3}} & \ps{\bestval{71.7}}{\bestval{50.0}} & \ps{69.7}{\runnerupval{41.7}} & \ps{49.8}{\bestval{25.0}} & \ps{\bestval{70.5}}{\bestval{45.8}} & 1.22 & \ct{43.2}{10.5} & \sfcell{90}{14} \\ 
\bottomrule
\end{tabular}
}
\par
\endgroup

\par\medskip
\begingroup
\normalsize
\centering
\caption{\textbf{Controlled interventions.} \modelGemini{}. P: Progress; S: Success (\%, $\uparrow$). Red superscripts: variant $-$ same-row Base. Bottom rows: gain $\Delta_v$ and gap change $I$ (points).}
\label{tab:intervention}
\renewcommand{\arraystretch}{1.12}
\setlength{\tabcolsep}{0.5pt}
\fontsize{8}{9.5}\selectfont
\definecolor{interventiondrop}{HTML}{A64B32}
\definecolor{interventiongain}{HTML}{16704A}
\newcommand{\ivbasecell}[2]{#1 & #2}
\newcommand{\ivhcell}[4]{\mbox{#1\textsuperscript{\fontsize{5}{5}\selectfont\textcolor{interventiondrop}{#3}}} & \mbox{#2\textsuperscript{\fontsize{5}{5}\selectfont\textcolor{interventiondrop}{#4}}}}
\newcommand{\ivvcell}[2]{\textcolor{interventiongain}{#1} & \textcolor{interventiongain}{#2}}
\newcommand{\ivicell}[2]{\textcolor{interventiongain}{#1} & #2}
\newlength{\interventionwidth}
\setlength{\interventionwidth}{\dimexpr\textwidth-24\tabcolsep\relax}
\resizebox{\textwidth}{!}{%
\begin{tabular}{@{}>{\raggedright\arraybackslash}p{0.17\interventionwidth} >{\centering\arraybackslash}p{0.085\interventionwidth} >{\centering\arraybackslash}p{0.09\interventionwidth} >{\centering\arraybackslash}p{0.085\interventionwidth} >{\centering\arraybackslash}p{0.065\interventionwidth} *{2}{>{\centering\arraybackslash}p{0.05\interventionwidth}} *{6}{>{\centering\arraybackslash}p{0.0675\interventionwidth}}@{}}
\toprule
\multirow{2}{*}{\textbf{Setting}}
& \multicolumn{4}{c}{\textbf{Assistance}}
& \multicolumn{2}{c}{Base} & \multicolumn{2}{c}{DA} & \multicolumn{2}{c}{TU} & \multicolumn{2}{c}{HC} \\
\cmidrule(lr){2-5}\cmidrule(lr){6-7}\cmidrule(lr){8-9}\cmidrule(lr){10-11}\cmidrule(lr){12-13}
& {\scriptsize Motion}
& {\scriptsize Attachment}
& {\scriptsize Grounding}
& {\scriptsize Map}
& P & S & P & S & P & S & P & S \\
\midrule
Motion-assisted & \yes & \no & \no & \no & \ivbasecell{80.0}{63.3} & \ivhcell{77.7}{53.3}{-2.3}{-10.0} & \ivhcell{73.7}{46.7}{-6.3}{-16.7} & \ivhcell{69.7}{43.3}{-10.3}{-20.0} \\
Friction-assisted & \no & \yes & \no & \no & \ivbasecell{80.7}{63.3} & \ivhcell{78.7}{56.7}{-2.0}{-6.7} & \ivhcell{74.7}{50.0}{-6.0}{-13.3} & \ivhcell{70.0}{46.7}{-10.7}{-16.7} \\
Grounding-assisted & \no & \no & \yes & \no & \ivbasecell{77.3}{58.6} & \ivhcell{75.0}{50.0}{-2.3}{-8.6} & \ivhcell{70.7}{46.7}{-6.6}{-12.0} & \ivhcell{66.7}{43.3}{-10.6}{-15.3} \\
Geometry-assisted & \no & \no & \no & \yes & \ivbasecell{79.7}{60.0} & \ivhcell{77.0}{53.3}{-2.7}{-6.7} & \ivhcell{73.0}{46.7}{-6.7}{-13.3} & \ivhcell{69.0}{46.7}{-10.7}{-13.3} \\
\addlinespace[1pt]
Fully assisted & \yes & \yes & \yes & \yes & \ivbasecell{82.0}{66.7} & \ivhcell{80.3}{60.0}{-1.7}{-6.7} & \ivhcell{76.0}{53.3}{-6.0}{-13.3} & \ivhcell{72.0}{50.0}{-10.0}{-16.7} \\
\midrule
\rowcolor{tableaccent}
\sys{} & \no & \no & \no & \no & \ivbasecell{76.0}{60.0} & \ivhcell{74.0}{50.0}{-2.0}{-10.0} & \ivhcell{69.3}{46.7}{-6.7}{-13.3} & \ivhcell{64.7}{43.3}{-11.3}{-16.7} \\
\midrule
\multicolumn{5}{@{}c}{\textit{$\Delta_v$: Fully assisted relative to \sys{}}}
& \ivvcell{+6.0}{+6.7} & \ivvcell{+6.3}{+10.0} & \ivvcell{+6.7}{+6.7} & \ivvcell{+7.3}{+6.7} \\
\multicolumn{5}{@{}c}{\textit{$I$: $\Delta_v(c)-\Delta_v(\mathrm{Base})$}}
& 0.0 & 0.0 & \ivvcell{+0.3}{+3.3} & \ivicell{+0.7}{0.0} & \ivicell{+1.3}{0.0} \\
\bottomrule
\end{tabular}
}
\par
\endgroup

\end{table}

\noindent\textbf{Mechanism ablations.}
\textit{\textbf{(1)} All six removals lower average Progress and Success.} Progress falls from 70.5\% to 51.0-67.7\%, and Success from 45.8\% to 20.8-41.7\% (\cref{tab:harness_ablation}).
Removing LiDAR and motion residuals lowers Success by 12.5 and 11.1 points, respectively; removing the grasp skill causes the largest decline, at 25.0 points.
\textit{\textbf{(2)} Local gains remain condition-specific.} Removing semantic spatial memory raises both PR and HC Success from 41.7\% to 58.3\%, but lowers average Progress and Success to 67.7\% and 41.7\%.
All variants retain current observations, task context, plans, dialogue, public events, and low-level primitives; a simpler context or interface may help individual episodes while weakening performance across the task mix.
The overall pattern supports combining perception, memory, and executable skills for task completion.

\noindent\textbf{Configuration trade-offs.}
\textit{\textbf{(1)} Configuration changes yield local gains.} Varying text history, image history, and actions per decision improves some metrics and conditions.
Two historical images improve Base Success from 58.3\% to 66.7\%, but average Progress falls from 70.5\% to 69.2\% and average Success to 38.9\%, while mean decisions increase from 43.2 to 47.0.
Removing text history reduces tokens per decision from 10.5k to 9.9k, yet requires 47.4 decisions and lowers Success to 36.1\%.
Four historical images reduce decisions to 41.3, with a lower Success of 37.5\%.
Longer text history and fixed three- or five-action horizons also use more tokens per decision without improving average Success.
\textit{\textbf{(2)} The default balances performance and cost:} ten text steps, the current image, and adaptive one-to-five-action decisions achieve the highest tested average Progress and Success at 43.2 decisions and 10.5k tokens per decision.

\subsubsection{Controlled Intervention Study}
\label{sec:controlled_interventions}

Using a dedicated sample of 30 fetch-and-deliver Base tasks and their DA, TU, and HC variants, we evaluate matched locomotion, attachment, grounding, and mapping interventions to assess whether execution assistance narrows capability gaps under unchanged task requirements.
For condition $c$ and assistance setting $a$, let $S_c^a$ denote Success in percent.
The within-setting capability gap and its change from normal execution are
\begin{equation}
G_c^a=S_c^a-S_{\mathrm{Base}}^a,
\qquad
I_c^a=G_c^a-G_c^{\mathrm{normal}}.
\label{eq:intervention_interaction}
\end{equation}
Positive $I_c^a$ narrows a negative Base-relative gap; the same definitions apply to Progress.
\Cref{tab:intervention} reports these descriptive contrasts for matched runs under the same task requirements and evaluation rules.

\Needspace{4\baselineskip}
\noindent\textbf{Execution gains do not imply smaller capability gaps.}
Full assistance improves Success across all conditions, but narrows the DA deficit by only 3.3 percentage points and leaves the TU/HC deficits unchanged at 13.3/16.7 points.
HC shows the largest Progress gain ($+7.3$) without a smaller Success gap relative to Base.
Thus, the tested execution assistance improves completion but does not eliminate sensitivity to task updates and human coordination.
Crossing execution interventions with capability variants reveals which deficits they reduce, extending evaluation beyond aggregate model rankings.

\begin{figure}[t]
\centering
\includegraphics[width=\textwidth]{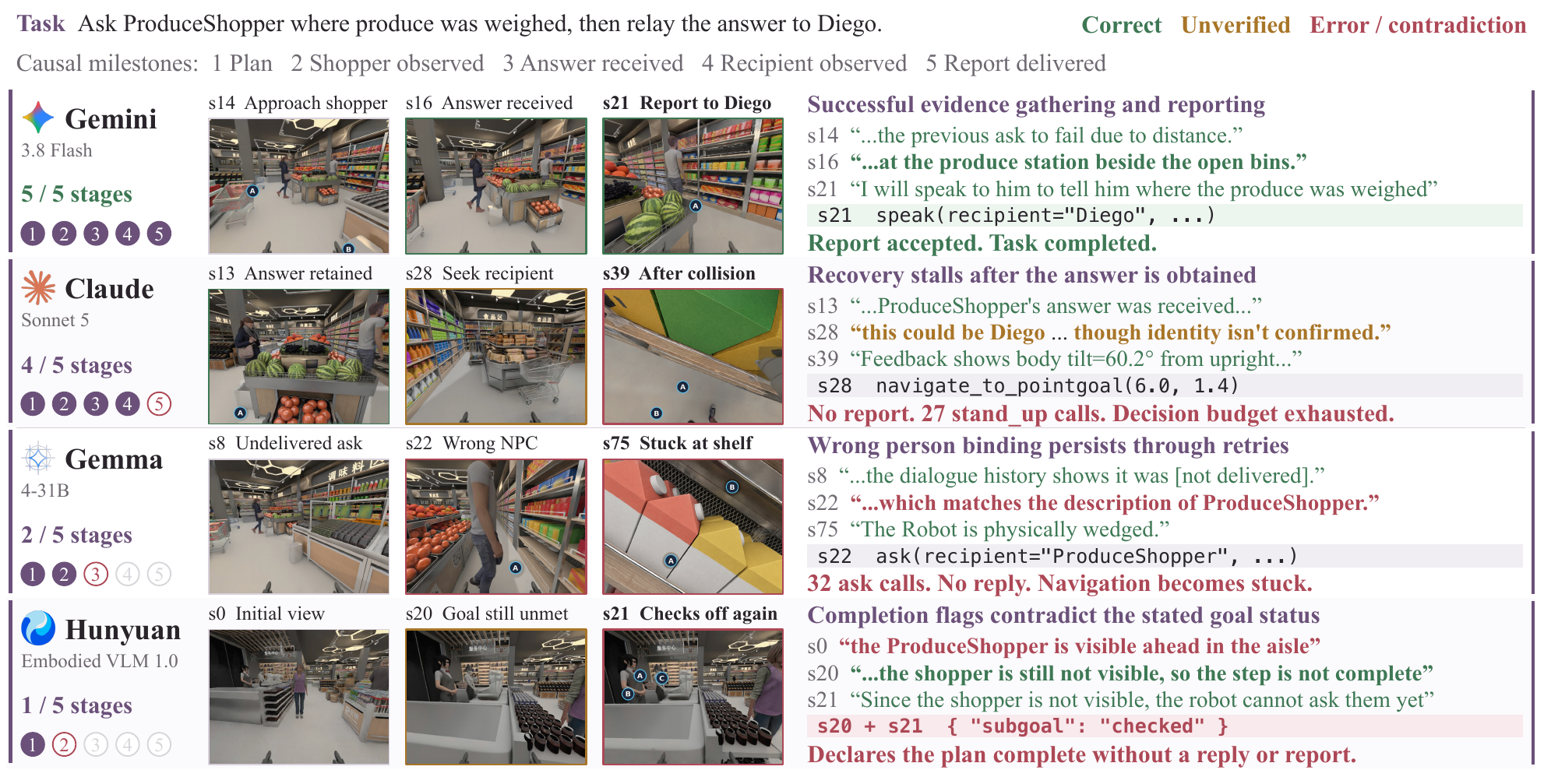}
\caption{\textbf{Qualitative comparison of four models on the same task.}\newline
Egocentric views, reasoning excerpts, and event-triggered causal milestones contrast successful reporting, recovery failure, wrong-person interaction, and unsupported completion claims.}
\label{fig:case}
\end{figure}
\FloatBarrier

\subsubsection{Failure Analysis}
\label{sec:failure_analysis}

\par\smallskip\noindent\begin{minipage}{\textwidth}
\begin{wrapfigure}{r}{0.46\textwidth}
\vspace{-\intextsep}
\setlength{\abovecaptionskip}{2pt}
\setlength{\belowcaptionskip}{0pt}
\centering
\includegraphics[width=\linewidth]{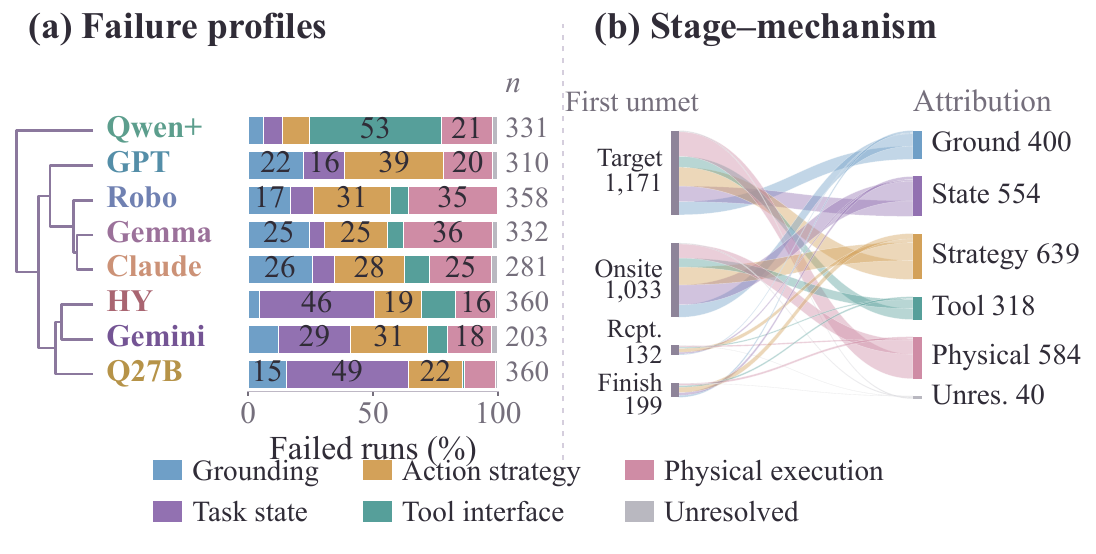}
\caption{\textbf{Failure analysis.}
\panelref{a} Model profiles. \panelref{b} Attributions by first unmet task outcome.}
\label{fig:failure_mechanisms}
\end{wrapfigure}

We distinguish the first unmet scored task outcome from the attributed failure mechanism.
\Cref{fig:failure_mechanisms} summarizes one primary failure-mechanism label for each of 2,535 failed runs in the binary-scored diagnostic dataset.
Labels come from model-assisted analysis of recorded trajectories and egocentric observations.
Tool-interface errors account for 52.9\% of \modelQwen{} failures; task-state errors account for 48.9\% and 46.1\% for \modelQwenOpen{} and \modelHyEmbodied{}, respectively.
Stopping stages locate unmet outcomes; trajectory evidence supports mechanism attribution.

For task updates, tracing revision delivery, goal revision, and physical completion distinguishes delivery from model-response failures.
\Cref{fig:case} compares four models on one \mainnav{} task.
\modelGemini{} completes all five stages, while \modelSonnet{} obtains the answer but exhausts its budget during recovery.
\modelGemma{} repeatedly addresses the wrong person, and \modelHyEmbodied{} marks its plan complete without obtaining or reporting the answer.

\finishwrap
\end{minipage}
\par\smallskip
\begin{table}[!t]
\captionsetup{skip=5pt}
\setlength{\abovecaptionskip}{5pt}
\setlength{\belowcaptionskip}{0pt}
\noindent\begin{minipage}[t]{0.48\textwidth}
\vspace{0pt}
\centering
\caption{\textbf{Skill calibration.}}
\label{tab:skill}
\renewcommand{\arraystretch}{1.10}
\setlength{\tabcolsep}{2pt}
\footnotesize
\begin{tabular*}{\linewidth}{@{\extracolsep{\fill}}l c c@{}}
\toprule
\multicolumn{1}{c}{\textbf{Setting}} & $\boldsymbol{N}$ & \textbf{Error $\downarrow$ / Success $\uparrow$} \\
\midrule
\compacttabgroup{3}{Locomotion}
\quad Translation (F/B/L/R) & \checkTranslationN & 1.3/3.1/4.9/4.7 cm \\[3.15pt]
\quad Rotation (L/R)        & \checkRotationN & 1.2/0.49$^{\circ}$ \\[3.15pt]
\quad Composite motion      & \checkCompositeN & 2.6 cm / 0.15$^{\circ}$ \\[3.15pt]
\addlinespace[1pt]
\compacttabgroup{3}{Grasping}
\quad Radius: 0.40 m  & \checkGraspN & \checkGraspNearPass/\checkGraspN \\[3.15pt]
\quad Radius: 0.475 m & \checkGraspN & \checkGraspMidPass/\checkGraspN \\[3.15pt]
\quad Radius: 0.55 m  & \checkGraspN & \checkGraspFarPass/\checkGraspN \\[3.15pt]
\quad Radius: 0.625 m & \checkGraspN & \checkGraspOuterPass/\checkGraspN \\[3.15pt]
\addlinespace[1pt]
\compacttabgroup{3}{Grounding}
\quad Detection & \checkGroundN & \checkDetectPass/\checkGroundN \\[3.15pt]
\quad Localization & \checkGroundN & \checkLocalizePass/\checkGroundN \\[3.15pt]
\bottomrule
\end{tabular*}

\end{minipage}\hfill
\begin{minipage}[t]{0.48\textwidth}
\vspace{0pt}
\centering
\caption{\textbf{Curation ablation.}}
\label{tab:curation}
\renewcommand{\arraystretch}{1.10}
\setlength{\tabcolsep}{2pt}
\footnotesize
\newlength{\curationwidth}
\setlength{\curationwidth}{\dimexpr\linewidth-4\tabcolsep\relax}
\begin{tabular}{@{}>{\raggedright\arraybackslash}p{0.60\curationwidth} C{0.10\curationwidth} C{0.30\curationwidth}@{}}
\toprule
\multicolumn{1}{c}{\textbf{Setting}} & $\boldsymbol{N}$ & \textbf{Usable (\%) $\uparrow$} \\
\midrule
w/o Multi-agent review       & \checkCurationN & \checkCurationAgent \\
w/o Human audit              & \checkCurationN & \checkCurationHuman \\
w/o \texttt{SKILL.md} refinement & \checkCurationN & \checkCurationSkill \\
w/o Simulation validation    & \checkCurationN & \checkCurationSimulation \\
\midrule
\rowcolor{tableaccent}
\bench{} curation            & \checkCurationN & \textbf{\checkCurationFull} \\
\bottomrule
\end{tabular}

\par\vspace{2pt}
\centering
\caption{\textbf{Human evaluation.}}
\label{tab:human_alignment}
\renewcommand{\arraystretch}{1.10}
\setlength{\tabcolsep}{2pt}
\footnotesize
\begin{tabular*}{\linewidth}{@{\extracolsep{\fill}}l c c c@{}}
\toprule
\textbf{Evaluation} & $\boldsymbol{N}$ & \textbf{Metric} & \textbf{Score $\uparrow$} \\
\midrule
Plan & \haPlanN & SRCC / PLCC & \haPlanRho\,/\,\haPlanPlcc \\
Inspect / Inquiry & \haReportN & SRCC / PLCC & \haReportRho\,/\,\haReportPlcc \\
\midrule
Causal & \number\numexpr\haHcTaskN+\haIuTaskN\relax & Pass (\%) & 100.0 \\
Feasibility & \number\numexpr\haHcTaskN+\haIuTaskN\relax & Pass (\%) & 100.0 \\
\bottomrule
\end{tabular*}

\end{minipage}
\end{table}

\subsubsection{Execution Checks}
\label{sec:execution_checks}

\cref{tab:skill} evaluates low-level execution under valid calls.
Translation errors range from 1.3 to 4.9\,cm; composite motion ends 2.6\,cm and $0.15^{\circ}$ from the target.
Grasping succeeds in \checkGraspTotalPass/\checkGraspTotalN{} trials, ranging from \checkGraspMinRate\% to \checkGraspMaxRate\% across four distances with \checkGraspN{} trials each.
\par
\begin{samepage}
\noindent
Grounding identifies \checkDetectPass{} correct instances and localizes \checkLocalizePass{} within \checkLocalizeTolerance\,cm of their 3D references across \checkGroundN{} visible-target views; detection failures count in both denominators.
Thus, recognition alone cannot ensure precise action.
Errors under valid calls motivate adapting subsequent actions to measured motion and contact feedback.
\par
\end{samepage}

\subsubsection{Curation Ablations}
\label{sec:curation_ablations}

\cref{tab:curation} ablates four components of \bench{} curation in \cref{fig:curation}(b--d), using \checkCurationN{} matched task specifications.
An independent audit checks task coherence and completion of required interactions and physical goals.
The full pipeline yields \checkCurationFull\% usable episodes; removing multi-agent review gives \checkCurationAgent\%, the largest drop of \checkCurationReviewDrop{} percentage points.
Removing human audit or \texttt{SKILL.md} refinement reduces yield to \checkCurationHuman\% or \checkCurationSkill\%, respectively; omitting simulation validation gives \checkCurationSimulation\%.
These comparisons support curator refinement alongside review and simulation.

\subsection{Human Evaluation}
\label{sec:human_alignment}

We compare semantic and human scores on \haEpisodeN{} trajectories and review \haHcTaskN{} HC and \haIuTaskN{} IU tasks for causal soundness and task feasibility.
\Cref{tab:human_alignment} reports SRCC/PLCC of \haPlanRho/\haPlanPlcc{} for plans and \haReportRho/\haReportPlcc{} for inspection/inquiry reports.
Causal and Feasibility report the percentages of reviewed tasks satisfying each criterion, respectively; both reach 100.0\% across the HC and IU tasks.

\section{Conclusion}
\label{sec:conclusion}

\sys{} grounds long-horizon robot tasks in shared physical and social constraints (\cref{sec:main_results,sec:execution_checks}).
\agentFramework{} couples robot dynamics, embodied NPC interaction, and state-machine control and evaluation; \agent{} lets frozen MLLMs act through memory, executable skills, and measured feedback (\cref{sec:harness_ablations,sec:human_alignment}).
\bench{} pairs two service task families with five capability demands, supporting trajectory-based comparisons of model decisions, completion, and execution quality as demands change during execution, including revised goals and dependencies on human participation.

\noindent\textbf{Limitations and Future Work.}
We evaluate one robot body and two indoor service task families in simulation.
Real-robot transfer must address differences in sensing, contact, and human behavior.
We plan Mobile-Franka transfer and task-adapted VLA, navigation, and manipulation backends.

\FloatBarrier
\bibliography{main}
\end{document}